\pdfoutput=1
\documentclass[11pt]{article}
\usepackage[dvipsnames]{xcolor}
\usepackage[]{emnlp2021}

\usepackage{times}
\usepackage{latexsym}

\newcommand{\bftab}{\fontseries{b}\selectfont}

\usepackage{xurl}

\usepackage[inline]{enumitem}
\usepackage{scrextend}
\usepackage{array}
\usepackage{arydshln}
\usepackage{graphicx}
\graphicspath{ {images/} }

\usepackage{hhline}

\usepackage{amsmath}

\usepackage{multirow}

\usepackage{enumitem}

\usepackage{arydshln}
\usepackage{array}
\newcolumntype{L}[1]{>{\raggedright\let\newline\\\arraybackslash\hspace{0pt}}m{#1}}
\newcolumntype{C}[1]{>{\centering\let\newline\\\arraybackslash\hspace{0pt}}m{#1}}
\newcolumntype{R}[1]{>{\raggedleft\let\newline\\\arraybackslash\hspace{0pt}}m{#1}}

\usepackage[T1]{fontenc}
\usepackage[utf8]{inputenc}

\usepackage{microtype}

\title{AStitchInLanguageModels: Dataset and Methods for the Exploration of Idiomaticity in Pre-Trained Language Models}

\author{
Harish Tayyar Madabushi, 
Edward Gow-Smith, \\
\textbf{Carolina Scarton} \and
\textbf{Aline Villavicencio}
\\[0.3cm]
Department of Computer Science, University of Sheffield \\
United Kingdom \\
\texttt{\small \{h.tayyarmadabushi, egow-smith1, c.scarton, a.villavicencio\} } \\[-0.15cm]
\texttt{\small @sheffield.ac.uk}
\\
}

\begin{document}
\maketitle
\begin{abstract}
Despite their success in a variety of NLP tasks, pre-trained language models, due to their heavy reliance on compositionality, fail in effectively capturing the meanings of multiword expressions (MWEs), especially idioms. Therefore, datasets and methods to improve the representation of MWEs are urgently needed. Existing datasets are limited to providing the degree of idiomaticity of expressions along with the literal and, where applicable, (a single) non-literal interpretation of MWEs. This work presents a novel dataset of naturally occurring sentences containing MWEs manually classified into a fine-grained set of meanings, spanning both English and Portuguese. We use this dataset in two tasks designed to test i) a language model’s ability to detect idiom usage, and ii) the effectiveness of a language model in generating representations of sentences containing idioms. Our experiments demonstrate that, on the task of detecting idiomatic usage, these models perform reasonably well in the one-shot and few-shot scenarios, but that there is significant scope for improvement in the zero-shot scenario. On the task of representing idiomaticity, we find that pre-training is not always effective, while fine-tuning could provide a sample efficient method of learning representations of sentences containing MWEs.
\end{abstract}

\section{Introduction and Motivation}

Pre-trained language models such as BERT~\cite{devlin-etal-2019-bert} and XLNet~\cite{NEURIPS2019_dc6a7e65} have been widely used in a variety of Natural Language Processing tasks. Despite their success in multiple downstream applications, such as sentence classification \cite{zhang-etal-2019-ernie} and reading comprehension~\cite{DBLP:journals/corr/abs-1910-10683}, they are unable to effectively represent idiomatic multiword expressions (MWEs) ~\cite{yu-ettinger-2020-assessing,garcia-etal-2021-probing}. Capturing idiomaticity is particularly challenging as the representations of words and phrases are explicitly designed to be compositional both in non-contextual \cite{mitchell2010composition,10.5555/2999792.2999959} and contextual embedding models. Pre-trained language models in particular exploit compositionality at both the word and sub-word levels~\cite{devlin-etal-2019-bert} to reduce the size of their vocabulary, which makes representing idiomatic phrases particularly challenging.
The effective representation of idiomatic MWEs is critical for them to be correctly interpreted in downstream tasks. Such an improvement will benefit both classification-based problems (e.g. sentiment analysis) and sequence-to-sequence tasks (e.g. machine translation).

To this end, we present a dataset consisting of naturally occurring sentences containing potentially idiomatic MWEs and two tasks aimed at evaluating language models' ability to effectively detect and represent idiomaticity. The primary contributions of this work are: 
\vspace{-0.25em}
\begin{enumerate}[topsep=5pt, partopsep=1pt,itemsep=1pt,parsep=1pt,leftmargin=12pt,listparindent=0pt]
    \item A novel dataset consisting of: 
    \begin{enumerate}
        \item naturally occurring sentences (and two surrounding sentences) containing potentially idiomatic MWEs annotated with a fine-grained set of meanings: compositional meaning, idiomatic meaning(s), proper noun and ``meta usage''; 
        \item paraphrases for each meaning of each MWE;
    \end{enumerate}
    \item Two tasks aiming at evaluating i) a model’s ability to detect idiomatic usage, and ii) the effectiveness of sentence embeddings in representing idiomaticity. \autoref{table:tasks} provides details of these tasks and associated subtasks, each designed to test different aspects of models. 
    \vspace{-0.2em}
    \begin{enumerate}
        \item These tasks are presented in multilingual, zero-shot, one-shot and few-shot settings. %to additionally evaluate the generalisability of models.
        \item We provide strong baselines using state-of-the-art models, including %the first tests of 
        experiments with one-shot and few-shot setups for idiomaticity detection and the use of the \emph{idiom principle} for detecting and representing MWEs in contextual embeddings. Our %experiments demonstrate 
        results highlight the significant scope for improvement.
    \end{enumerate}
\end{enumerate}

\begin{table}[ht]
\footnotesize
\begin{center}
\renewcommand{\arraystretch}{1.25} 
\begin{tabular}{|L{0.2cm}|L{1.3cm}|L{4.8cm}|} 
\hline 
\multirow{2}{*}{\rotatebox{90}{Task 1    }} & Subtask A & Coarse-grained classification of examples containing idioms.  \\
                                            & Subtask B & Fine-grained classification of examples into  meanings. \\
\hdashline 
\multirow{2}{*}{\rotatebox{90}{Task 2    }} & Subtask A & Effective representation of sentences containing idiomatic phrases using only \emph{pre-training}.  \\
                                            & Subtask B & Effective representation of sentences using both pre-training and \emph{fine-tuning}. \\
\hline
\end{tabular}
\end{center}
\caption{\label{table:tasks} AStitchInLanguageModels Tasks: The two tasks and associated subtasks.}
\end{table}

This dataset and associated tasks have the potential to catalyse research into representing more complex elements of language beginning with idiomaticity, thus ensuring a timely \textit{stitch in language models}. We call this dataset and associated tasks \emph{AStitchInLanguageModels}, and make the dataset, the associated splits for each task, pre-training data, pre-trained and fine-tuned models, program code and associated processing scripts, including hyperparameters, publicly available in the interest of reproducibility and for subsequent reuse\footnote{\footnotesize	\url{https://github.com/H-TayyarMadabushi/AStitchInLanguageModels}}. 

This paper is organised as follows: Section \ref{sec:related} presents a discussion of related work. We then present AStitchInLanguageModels consisting of the novel MWE dataset and the two associated tasks in Section \ref{section:dataset}. We discuss our experiments and results for these two tasks in Section \ref{section:experiments-and-results}, before presenting a discussion of the more interesting elements of our findings in Section \ref{section:discussion}. We present our conclusions and possible avenues of future work in Section \ref{section:conclusions-and-future-work}. 

\section{Related work}
\label{sec:related}

The problems posed by MWEs to NLP models have been known for some time~\cite{10.5555/647344.724004,constant-etal-2017-survey,shwartz-dagan-2019-still}. For instance,
~\newcite{10.5555/647344.724004} refer to the \emph{idiomaticity problem} and place the need for effective processing of MWEs on par with that for word sense disambiguation to be able to effectively process text. While their analysis focused on symbolic
%, as opposed to statistical, 
methods, this problem still persists:
%in the paradigm of statistical methods using distributional semantics (e.g. ~\newcite{mitchell2010composition,reddy-etal-2011-empirical}). More recently, 
\newcite{shwartz-dagan-2019-still} showed, using six tasks, that %the introduction of 
contextual pre-trained language models, capable of handling polysemy, continued to be unable to effectively handle idiomatic MWEs, although they tend to do better than their non-contextual predecessors. Further experiments with probing pre-trained language models across multiple languages have also confirmed this result~\cite{yu-ettinger-2020-assessing,garcia-etal-2021-probing}.

\subsection{Existing Datasets}
\label{section:existing-datasets}
Datasets of MWE annotated corpora include that associated with the PARSEME shared task~\cite{savary-etal-2017-parseme} which focuses on verbal MWEs and the STREUSLE dataset~\cite{schneider-etal-2014-comprehensive,schneider-smith-2015-corpus,schneider-etal-2016-corpus} which includes noun, verb, prepositional and possessive expressions including ``semantic supersenses''. However, most existing datasets associated with compositionality of MWEs consist of isolated phrases, labelled with overall compositionality scores~\cite{venkatapathy-joshi-2005-measuring,biemann-giesbrecht-2011-distributional,farahmand-etal-2015-multiword}, scores of how individual words contribute to the meaning of the MWE~\cite{venkatapathy-joshi-2005-measuring}, or both~\cite{reddy-etal-2011-empirical,cordeiro-etal-2019-unsupervised,schulte-im-walde-etal-2016-ghost}. While most of these target only English, some include scores for other languages such as German \cite{schulte-im-walde-etal-2016-ghost}, and French and Portuguese~\cite{cordeiro-etal-2019-unsupervised}. 

Existing datasets of compositionality that include context often add context automatically by first selecting MWEs that are either only compositional or only idiomatic. For instance, the VNC-Tokens Dataset~\cite{Cook08thevnctokens} consists of 53 English MWEs each with a maximum of 100 sentences extracted from the BNC, while ~\newcite{tu-roth-2012-sorting} collected 1,348 sentences associated with 23 verb phrases annotated as compositional and idiomatic.~\newcite{shwartz-dagan-2019-still} focused on a subset of noun compounds that are only compositional or idiomatic from the dataset provided by ~\newcite{reddy-etal-2011-empirical} and automatically added sentences from Wikipedia. Finally, the NCS Dataset~\cite{garcia-etal-2021-probing} consists of 280 English and 180 Portuguese MWEs, annotated with degrees of compositionality and three sentences containing each of the MWEs.

Despite the importance of the context surrounding an MWE, where available, context, in the form of sentences containing MWEs, is available only for those MWEs that are either idiomatic or compositional. This significant shortcoming makes it impossible to train models to learn to differentiate between the compositional and idiomatic usage of the same MWE. 

Finally, while existing datasets also provide paraphrases for the compositional and idiomatic meanings of MWEs~\cite{hendrickx-etal-2013-semeval,garcia-etal-2021-probing}, they are limited to having exactly one compositional and one idiomatic meaning, which is not always the case as is exemplified by the phrase ``head hunter'' which, while not having a literal usage, has multiple idiomatic meanings (i.e recruiter, baseball pitcher who aims for the head, and hunter). 

\emph{AStitchInLanguageModels} is designed to alleviate these shortcomings, specifically: a) the lack of context sentences, b) the need for fine grained classification of MWEs, and a more complete set of paraphrases for all possible meanings of MWEs (Section \ref{section:dataset}). 

\subsection{Methods}
\label{section:existing-methods}
The task of identifying idiomaticity in sentences was initially addressed by use of symbolic methods~\cite{baldwin-villavicencio-2002-extracting,10.5555/647344.724004}, statistical properties of text such as mutual information~\cite{lin-1999-automatic}, and latent semantic analysis~\cite{baldwin-etal-2003-empirical}.

The subsequent adoption of distributional semantics led to the use of constituent word embeddings to determine the compositionality of phrases, such as in the work by ~\newcite{katz-giesbrecht-2006-automatic} who made use of the semantic similarity between the distributional vectors associated with an MWE as a whole and those associated with its parts to determine compositionality. This is achieved by use of a single token to represent an MWE. This trend continued with the introduction of neural distributional semantic models such as word2vec~\cite{word2vec} wherein MWEs were taken as single units in learning embeddings~\cite{10.5555/2999792.2999959}. This method was improved upon by use of an explicit disambiguation step prior to composition~\cite{kartsaklis-etal-2014-resolving}, and by the joint learning of compositional and idiomatic embeddings using a ``compositionality scoring'' function~\cite{hashimoto-tsuruoka-2016-adaptive}. This ``single token'' method has the advantage of being rooted in the linguistic \emph{idiom principle}~\cite{sinclair1991corpus}, which postulates that humans process idioms by treating them as a ``single independent token''.

Despite being the only method of handling MWEs and having had relative success, it is not without its shortcomings. The first is that the frequency of MWEs tends to be low (a problem that worsens with the increase in length of MWEs) and since the quality of distributional representations tends be proportional to the number of instances of a token, representations of MWEs are often lacking. The second is that non-contextual type level representations are inherently limited as MWEs often have multiple meanings, as detailed in Section \ref{section:existing-datasets}. 

While contextual embeddings can handle polysemy, they fail to fully capture the meaning of MWEs as discussed earlier. How contextual embeddings fair in comparison to their non-contextual predecessors is not entirely clear as~\newcite{nandakumar-etal-2019-well} found that they do worse on some tasks while~\newcite{shwartz-dagan-2019-still} found that they do better.~\newcite{hashempour-villavicencio-2020-leveraging} adopted the idiom principle (MWE as a single token) with contextual language models (specifically BERT), and found that this method does not benefit transformer-based pre-trained models. However, they did not introduce a new token to represent each MWE as is required during the training of non-contextual models built on the idiom principle, but instead replaced MWEs with a single token in the input and rely on BERT's  word-piece tokenizer. To the best of our knowledge this work is the first to introduce new tokens for MWEs into a contextual pre-trained language model (see Section \ref{section:experiments-results-task2}).

\section{AStitchInLanguageModels: Dataset and Tasks} 
\label{section:dataset}
To create a dataset and tasks aimed at improving language models' ability to identify and capture idiomaticity, we first collected examples of MWE usage in naturally occurring sentences along with the two surrounding sentences. We then annotated these examples with a fine-grained set of meanings associated with each usage. We restrict our attention to noun compounds, a subset of idiomatic MWEs, sourced from the Noun Compound Senses (NCS) dataset~\cite{cordeiro-etal-2019-unsupervised}, which extends the dataset by~\newcite{reddy-etal-2011-empirical}.

\subsection{Data Collection and Annotation}

A total of 12 judges were asked to collect examples containing a list of MWEs occurring naturally in context, in both English and Portuguese. For each MWE, judges were instructed to obtain 7 to 10 examples of each meaning ( ``Idiomatic'',  ``Non-Idiomatic'',  ``Proper Noun'' and  ``Meta Usage'') where possible, with between 20 and 30 total examples for each MWE. We define ``Meta Usage'' to be the literal use of an MWE in a metaphor (e.g. life vest in ``Let the Word of God be our life vest to keep us afloat, so as not to drown.''). Judges were additionally instructed to add to the list of possible meanings associated with each MWE based on the usage they observed when collecting examples, or to flag examples with novel usage for review by language experts. Emphasis was put on extracting high-quality examples with three contiguous sentences and correct formatting, containing no unusual characters. The data consists of excerpts of text from the web, each a maximum of three sentences, thus adhering to fair use.

The meanings of each MWE were then paraphrased by language experts. The idiomatic paraphrases aim to concisely convey the meaning of the idiom. For example, \emph{cutting edge} is paraphrased to \emph{most advanced} and \emph{night owl} is paraphrased to \emph{nocturnal person} in the idiomatic case. The aim of the literal paraphrase is to apply a minimal lexical alteration that shifts the MWE away from its idiomatic meaning(s). For example, \emph{cutting edge} is paraphrased to \emph{slicing edge} and \emph{night owl} is paraphrased to \emph{night hooter} in the literal case. This adversarial paraphrasing is designed to test a model's ability to discern a compositional meaning from an idiomatic one, and aims to ensure that models must have a nuanced understanding of idiomaticity for them to be successful.
Examples of the annotated data are shown in ~\autoref{table:datasetExamples}.

\begin{table*}[ht]
\centering\resizebox{\textwidth}{!}{%
\begin{tabular}{ |c | p{5cm} | p{3cm}| p{3cm}| c | c| c |} 
\hline
MWE & Target Sentence & Previous Sentence & Next Sentence & Label & Idiomatic? & Paraphrase \\ 
\hline
gold mine & This means that search data is a \emph{gold mine} for marketing strategy. (\href{https://www.marketingweek.com/search-data-gold-mine-marketing-strategy/
}{marketingweek.com})  & The data that those searches generate builds... & It reveals which types of product... & Idiomatic 1 & Yes & source of fortune \\
\hline
gold mine & The hashtag “Qixia \emph{gold mine} incident” has been viewed many million of times on the social media site Weibo. (\href{https://www.wsws.org/en/articles/2021/01/26/chin-j26.html}{wsws.org}) &The rescue operation took place... & A week after the explosion... & Literal & No & mine \\
\hline
gold mine & The \emph{Gold Mine}’s plain frontage \& sparse, white-walled dining room suggest that it’s a quick-fix refuelling stop rather than a place to linger. (\href{https://www.squaremeal.co.uk/restaurants/gold-mine_1601}{squaremeal.co.uk}) & SquareMeal Review of Gold Mine & The menu touts a bewildering array of dishes... & Proper Noun & No & Proper Noun \\
\hline 
\end{tabular}}
\caption{\label{table:datasetExamples}A sample of the dataset for one MWE (\emph{gold mine}). Context sentences are truncated for brevity. The "Idiomatic?" column is used for the coarse-grained classification task (Subtask 1 A) and the "Paraphrase" column is used for the fine-grained classification task (Subtask 1 B) and the representation task (Task 2). }
\end{table*}

Finally, each example was annotated with a label and corresponding paraphrase by two judges. The Cohen's kappa coefficient of inter-rater reliability was 0.887 for English and 0.807 for Portuguese. We note that a significant proportion of disagreements arose from a difference in interpretation of the ``Proper Noun'' and ``Meta Usage'' labels, and from what constituted ``low quality'' for discarding examples. For resolution of disagreements a final label was decided based on a discussion between the judges.

\subsection{The Final Dataset}
The final dataset consists of 4,558 English examples containing 223 MWEs, and 1,872 Portuguese examples containing 113 MWEs. 

We divide this data into training, development and test splits as follows: the test and development splits consist of sentences containing 30 and 20 idioms each in English and Portuguese respectively. To enable the testing of models under different scenarios of data availability, we create three different setups of the test split for each language. The first, the \emph{zero-shot} setup, consists of sentences containing 163 idioms in English and 60 idioms in Portuguese, which do not occur in the development and test sets. The second, the \emph{one-shot} setup, consists of exactly one non-idiomatic and (where available) one idiomatic example associated with each MWE in the development and test sets. The third and final, the \emph{few-shot} setup, consists of between 1 and 4 examples associated with each meaning of each MWE in the development and test sets. The exact number of examples available is proportional to the original number of examples associated with that specific meaning of that idiom. We make it clear that there are no overlapping target sentences between the three splits - the only overlap is in terms of the idioms contained in examples. Detailed statistics for the English and Portuguese datasets are provided in \autoref{app:app-datasetstatistics}.

\subsection{Tasks}
In addition to the dataset of labelled contextualised MWEs, we present two tasks.

\subsubsection{Task 1: Idiomaticity Detection}
The first task we propose is designed to evaluate the extent to which models can identify idiomaticity in text and consists of two Subtasks: a \emph{coarse-grained} classification task (Subtask A) and a \emph{fine-grained} classification task (Subtask B). For the coarse-grained subtask, the problem is simplified to classifying the examples as either ``Idiomatic'' or ``Non-Idiomatic''. For the purposes of this subtask, anything labelled as ``Literal'' or ``Proper Noun'' was classed as ``Non-Idiomatic'' and given a label of 1, whilst all ``Idiomatic'' labels as well as ``Meta Usage''  were given a label of 0. (See also \autoref{table:datasetExamples}).

For the fine-grained task, the possible meanings are equivalent to the paraphrases in the dataset, described previously. Since this problem does not have a fixed number of labels (given that each MWE has a different set of meanings), we convert this to a binary classification problem: the first input is the example containing the MWE and the second the paraphrase of each possible meaning of the MWE (or one of the phrases ``Proper Noun'' or ``Meta Usage''). An input pair is labelled 1 if the paraphrase represents the correct meaning of the MWE in the example and 0 otherwise. In addition, we report scores for both subtasks in the zero-shot, one-shot and few-shot setups %of the problem so as 
to better evaluate a model's ability to generalise and learn in a sample efficient fashion. We note that this was impossible prior to the introduction of AStitchInLanguageModels as all previous datasets which included context considered only one meaning per MWE (see Section \ref{section:existing-datasets}). Due to the imbalanced nature of these subtasks, we use Macro F1-score as the measure of evaluation. 

We note that due to the different ways in which the two settings in this Task are setup the results for the two settings in this task are not directly comparable. 

\begin{table*}[ht]
\footnotesize
\def\arraystretch{1.1}
\centering
\begin{tabular}{|L{3.8cm}|L{3.8cm}|L{3.8cm}|C{2.9cm}|}
\hline
Sentence (E) & Correct Replacement (\scalebox{0.8}{$E_{\text{MWE} \rightarrow \text{c}}$}) & Wrong Replacement (\scalebox{0.8}{$ E_{ \text{MWE}\rightarrow \text{i}}$}) & Expected \\
\hline
When removing a \emph{big fish} from a net, it should be held in a manner that supports the girth. (\href{https://www.newsdakota.com/2021/03/15/our-outdoors-big-fish-picture-tips/}{newsdakota.com}) & When removing a \emph{fish} from a net, it should be held in a manner that supports the girth. & When removing a \emph{important person} from a net, it should be held in a manner that supports the girth. & \scalebox{0.7}{ 
\begin{tabular}{@{}l@{}}
  $sim(E, E_{\rightarrow \text{c}}) = 1$ \\ 
  $sim(E, E_{\rightarrow \text{i}}) = sim(E_{\rightarrow \text{c}}, E_{\rightarrow \text{i}})$
 \end{tabular}}\\
\hdashline
To pay attention only to new housing and houses I think skews the \emph{big picture}. (\href{https://streets.mn/2021/03/16/build-more-housing-the-big-picture/}{streets.mn}) & To pay attention only to new housing and houses I think skews the \emph{whole situation}. &
To pay attention only to new housing and houses I think skews the \emph{large image}.  &  \scalebox{0.7}{ 
\begin{tabular}{@{}l@{}}
  $sim(E, E_{\rightarrow \text{c}}) = 1$ \\ 
  $sim(E, E_{\rightarrow \text{i}}) = sim(E_{\rightarrow \text{c}}, E_{\rightarrow \text{i}})$
 \end{tabular}}\\
\hline
\end{tabular}
\caption{\label{table:subtaskc-v2} Task 2 - Models are required to be consistent in assigning semantic similarity scores as measured by use of the paraphrases of different meanings.}
\end{table*}

\subsubsection{Task 2: Idiomaticity Representation} \label{section:task2-intro}
While the identification of idiomaticity is important, downstream tasks require embeddings that effectively capture idiomaticity, which is the purpose of the second task. For this task, we design a metric to measure how consistent a model is in capturing similarity between sentences containing idiomatic elements and sentences that are purely compositional. 

As each possible meaning of an MWE contained in each example is associated with a paraphrase, this task requires a model to generate similarity scores for each example $E$ such that: 

\begin{equation}
\label{equation:task2}
\begin{split}
    \forall_{i \in I} \Big(  &sim(E, E_{\rightarrow \text{c}}) = 1; \\
    & sim(E, E_{\rightarrow \text{i}}) = sim(E_{\rightarrow \text{c}}, E_{\rightarrow \text{i}}) \Big)
\end{split}
\end{equation}

\noindent where $E_{\rightarrow \text{c}}$ represents the example with the MWE in $E$ replaced by the paraphrase of the correct meaning associated with the MWE, and $E_{\rightarrow \text{i}}$ the example with the MWE replaced by a paraphrase of one of the incorrect meanings of the MWE in $E$~(see \autoref{table:subtaskc-v2} for examples).

Without additional checks, models can trivially succeed in this task by simply assigning a similarity score of 1 to every sentence pair. To prevent this, we splice in development and test data from the Semantic Text Similarity (STS) benchmark dataset~\cite{cer-etal-2017-semeval} in English and from the ASSIN2 STS dataset~\cite{real2020assin} for Portuguese. %% 500 for both (ignore what readme says)

We note that the expected similarity scores are approximates as the paraphrases need not have \emph{exactly} the same meanings as that of the MWE they are paraphrasing. However, we consider this difference to be acceptable given the typical nature of annotation of semantic similarity data wherein annotators use labels between 1 and 5.  

Finally, we divide this task into two subtasks: Subtask A which requires the solving of this task using \emph{only pre-training} and Subtask B which allows the \emph{fine-tuning} of models. For clarity, we define pre-training to be the training of a model on any task \emph{other than idiomatic STS} (and can include ``fine-tuning'' on a different task), and fine-tuning to include the inclusion of training on any STS dataset which includes potentially idiomatic MWEs. We use Spearman correlation coefficient as the measure of evaluation for both subtasks in Task 2 as it has been shown that Pearson correlation is poorly suited for comparing performance on the STS task~\cite{reimers-etal-2016-task}. 

\section{Experiments and Results}
\label{section:experiments-and-results}
Our aim was to investigate the performance of state-of-the-art transformer-based pre-trained language models on these tasks, and how their performance varied with different input features (i.e. inclusion of MWE, context sentences), problem setups (i.e. zero-shot, one-shot and few-shot), and training regimes (i.e. pre-training, fine-tuning) so as to provide a baseline for the AStitchInLanguageModels dataset.
Here we provide an overview of the experiments ran and our results. More detailed description of the experimental procedure, including runtimes are given in \autoref{app:app-experimentalProcedure}.

\subsection{Task 1: Idiomaticity Detection}
\label{sec:experimentsresults-task1}

For Subtask A, which requires the
coarse-grained classification of examples, we start by exploring the impact of three variables in the zero-shot setup: the pre-trained language model, the inclusion of context (the two surrounding sentences), and adding
the relevant MWE as a feature. The context is included by simply concatenating the three contiguous sentences, and the MWE is included by separating it from the rest of the input by use of the ``[SEP]'' token. 
For the purposes of brevity, we report a subset of variations highlighting the most interesting results, with more details of the experimental procedure in  \autoref{app:app-experimentalProcedure}. Among the results for Task 1, Subtask A (\autoref{table:resultsTask1a}), the best-performing experimental settings from the zero-shot setting (by development F1 score) were transferred over to the one-shot and few-shot problem setups.  
While the inclusion of context (surrounding sentences) did not change the performance of the models significantly, and will not be used in the other experiments, the inclusion of the relevant MWE was found to be beneficial to performance. 

For Subtask B, fine-grained classification, the best-performing experimental settings found for the first subtask were used for the multiclass data, although the MWE was not included as a feature, since our previous method for inclusion is incompatible with the passing of the paraphrase; the input consists of the target sentence without the previous or next sentences followed by a single possible meaning of the MWE separated by the ``[SEP]'' token. The task is thus reduced to a binary classification task wherein the model is required to predict 1 when the target sentence is followed by the correct paraphrase and 0 otherwise. The results are in ~\autoref{table:resultsTask1b}. 

\begin{table*}[ht]
\centering
\footnotesize
\begin{tabular}{ | c | c|c c c  | c c |} 
\hline
& Problem Setup & Model & Context? & MWE? & Dev F1 & Test F1 \\
\hline
\multirow{6}{*}{\rotatebox{90}{English~~~~~~~~~~~~~~~~~~}} & \multirow{4}{*}{zero-shot} & BERT base (cased) & No & No & 0.724 & 0.688 \\ 
&& BERT base (cased) & Yes & No & 0.717 & 0.797 \\ 
&& BERT base (cased) & Yes & Yes & 0.779 & 0.774\\ 
 && BERT base (cased) & No & Yes & 0.785 & 0.821 \\ 
 && XLNet base (cased) & No & Yes & \bftab 0.823 & \bftab 0.832 \\ 
\cline{2-7}
& one-shot & XLNet base (cased) & No & Yes & \bftab 0.897 & \bftab 0.874\\ 
& one-shot & XLNet base (cased) & Yes & No & 0.689 & 0.701\\ 
& one-shot & XLNet base (cased) & No & No & 0.755 & 0.754\\
\cline{2-7}
&few-shot & XLNet base (cased) & No & Yes & \bftab 0.959 & \bftab 0.971\\ 
&few-shot & XLNet base (cased) & Yes & No & 0.782 & 0.806\\
&few-shot & XLNet base (cased) & No  & No & 0.792 & 0.853\\
\hline
\multirow{6}{*}{\rotatebox{90}{Portuguese~~~~~~~~~~~~~~~~}} & \multirow{4}{*}{zero-shot} & XLM-RoBERTa base (cased) & No & No & 0.593 & 0.528 \\ 
 && XLM-RoBERTa base (cased) & Yes & No & 0.542 & 0.562 \\
 && XLM-RoBERTa base (cased) & Yes & Yes & 0.696 & \bftab 0.604 \\
 && XLM-RoBERTa base (cased) & No & Yes & \bftab 0.703 & 0.579 \\ 
 && BERT base multilingual (cased)& No & Yes & 0.686 & 0.560 \\ 
\cline{2-7}
& one-shot & XLM-RoBERTa base (cased) & No & Yes & \bftab 0.877 & \bftab 0.778\\ 
& one-shot & XLM-RoBERTa base (cased) & Yes & No & 0.605 & 0.563\\
& one-shot & XLM-RoBERTa base (cased) & No  & No & 0.638 & 0.534\\
\cline{2-7}
&few-shot & XLM-RoBERTa base (cased) & No & Yes & \bftab 0.926 & \bftab 0.944 \\
&few-shot & XLM-RoBERTa base (cased) & Yes & No & 0.655 & 0.684\\
&few-shot & XLM-RoBERTa base (cased) & No  & No & 0.796 & 0.696\\
\hline
\end{tabular}
\caption{\label{table:resultsTask1a}Evaluation results for Task 1 Subtask A (with best results for each setting in bold).}
\end{table*}

\begin{table*}[ht]
\centering
\footnotesize
\begin{tabular}{ | c | c | c | c c |} 
\hline
& Problem Setup & Model & Dev F1 & Test F1\\
\hline
\multirow{3}{*}{\rotatebox{90}{En}} & zero-shot & XLNet base (cased) & 0.852 & 0.875 \\ 
\cline{2-5}
& one-shot & XLNet base (cased) & 0.923 & 0.927 \\ 
\cline{2-5}
&few-shot & XLNet base (cased) & 0.933 & 0.948 \\ 
\hline
\multirow{3}{*}{\rotatebox{90}{Pt}} & zero-shot & XLM-RoBERTa base (cased) & 0.843 & 0.778 \\ 
\cline{2-5}
& one-shot & XLM-RoBERTa base (cased) & 0.852 & 0.858  \\ 
\cline{2-5}
& few-shot & XLM-RoBERTa base (cased) & 0.909 & 0.878\\ 
\hline
\end{tabular}
\caption{\label{table:resultsTask1b}Evaluation results for Task 1 Subtask B.}
\end{table*}

\subsection{Task 2: Idiomaticity Representation}
\label{section:experiments-results-task2}
Task 2 requires models to output the semantic similarity between sentences in a consistent manner. Given that sentence embeddings generated by pre-trained language models cannot directly be used to calculate semantic similarity~\cite{devlin-etal-2019-bert}, we used Sentence BERT~\cite{reimers-2019-sentence-bert} which consists of a siamese network structure with a regression objective function consisting of the mean-squared error loss calculated over the cosine similarity of two input sentences during training. This results in sentences whose semantic similarity can be compared using cosine similarity~\cite{7298682}. We note that while this is not strictly required for our purpose, we use this method as the siamese network structure is likely to be beneficial in fine-tuning on the idiomatic STS data where the similarity scores are all relatively close to each other.

To test the effectiveness of the idiom principle to represent MWEs (Section \ref{section:existing-methods}) for Task 2, we analyse three different settings, involving the expansion of the vocabulary of pre-trained models 
by the addition of a single token to represent each MWE: In the first setting (``all replace'') all instances of an MWE are replaced with the corresponding token before input to the model; in the second (``select replace'') each input sentence is first classified using the one-shot model for course grained classification (Section \ref{sec:experimentsresults-task1}) and a given instance of an MWE is replaced only when the one-shot model predicts that the MWE in a given sentence has an idiomatic meaning; and in the third (``no replace'') there is no change to either the model (no special token added) or their input. 

For Subtask A, which requires the use of only pre-training, we collect sentences (including, where available, the paragraph they occur in) from the Common Crawl News Dataset\footnote{\href{https://commoncrawl.org/2016/10/news-dataset-available/}{https://commoncrawl.org}} spanning the first 6 months of 2020 (over half a terabyte of text). This results in about 220,000 sentences in English and about 16,000 in Portuguese containing relevant MWEs. We use this data to continue pre-training BERT base in both the ``all replace'' and ``select replace'' variations described above. Unlike our other experiments, we do not pre-train multiple times due to time and resource constraints. We also limit pre-training to 5 epochs for English and 10 epochs for Portuguese based on results from our exploratory experiments. 

In addition to these two models, we also test BERT base with no modifications, and a version of BERT base with the addition of tokens associated with each MWE but no pre-training (the embeddings associated with these tokens are randomly initialised). The ``all replace'' and ``select replace'' models have their pre-training and input sentences tokenized according to the same strategy. Each of these models are subsequently trained using the Sentence BERT architecture so as to ensure that the resultant embeddings can be compared using cosine similarity. We train using the training data from the STS benchmark dataset~\cite{cer-etal-2017-semeval} for English and the ASSIN2 STS dataset~\cite{real2020assin} for Portuguese. This training does \emph{not} violate the ``pre-train only'' requirement of this task as we do not train on idiomatic STS data. The results are presented in \autoref{table:task2-a-results}. 

\begin{table}[ht]
\footnotesize
\def\arraystretch{1.1}
\centering
\begin{tabular}{|L{0.3cm}|L{2.5cm}|L{0.75cm}|L{1cm}|}
\hline
& Tokenization & Dev $\rho$ & Test $\rho$ \\
\hline
\multirow{4}{*}{\rotatebox{90}{English~~~~~~}} & Default & 0.767 & 0.744 \\
& All Tokenized (No Pre-Training) & 0.826 & 0.801 \\
& All Tokenized  & 0.835 & 0.811 \\
& Select Tokenized  & 0.848 & 0.805 \\
\hline
\multirow{4}{*}{\rotatebox{90}{Portuguese~~~}} & Default & 0.726 & 0.785\\
& All Tokenized (No Pre-Training) & 0.749 & 0.798 \\
& All Tokenized  & 0.742 & 0.805 \\
& Select Tokenized  & 0.750 & 0.814 \\
\hline
\end{tabular}
\caption{\label{table:task2-a-results} Results for Task 2 Subtask A. }
\end{table}

For Subtask B, we fine-tune the ``no replace'', ``all replace'' and ``select replace'' versions of BERT base on both the standard STS data as in Subtask A and training data constructed from the zero-shot and few-shot version of the training data using \autoref{equation:task2}. Therefore, during fine-tuning, the gold similarity score for $sim(E, E_{\rightarrow \text{c}})$ is $1$ and that for $sim(E, E_{\rightarrow \text{i}})$ is $sim(E_{\rightarrow \text{c}}, E_{\rightarrow \text{i}})$.
Both the ``replace'' versions of the model are fine-tuned from scratch (i.e. the tokens associated with MWE are random and not pre-trained as in Subtask A). Although it is possible to start with the pre-trained version of the ``replace'' models, we make the conscious decision not to, so we might test if this sample efficient method of learning is feasible. The results for these models are presented in \autoref{table:task2-b-results}.

\begin{table}[ht]
\footnotesize
\def\arraystretch{1.1}
\centering
\begin{tabular}{|L{0.3cm}|L{2.5cm}|L{1cm}|C{1cm}|}
\hline
& Tokenization & Dev $\rho$ & Test $\rho$ \\
\hline
\multirow{3}{*}{\rotatebox{90}{En }} & Default & 0.818 & 0.823 \\
& All Tokenized  & 0.821 & 0.817 \\
& Select Tokenized  & 0.851 & 0.825 \\
\hline
\multirow{3}{*}{\rotatebox{90}{Pt  }} & Default & 0.752 & 0.811 \\
& All Tokenized  & 0.803 & 0.835 \\
& Select Tokenized  & 0.806 & 0.818 \\
\hline
\end{tabular}
\caption{\label{table:task2-b-results} Results for Task 2 Subtask B.}
\end{table}

\section{Discussion}
\label{section:discussion}

This section discusses some highlights of our results.

\subsection{Detection of Idiomaticity}

In the task of detecting idiomaticity (Task 1), we find that in the zero-shot setting, the models perform poorly in both the coarse-grained and fine-grained subtasks. 
%Due to the imbalance in the test sets arising from the natural imbalance of MWE usage (see \autoref{app:app-datasetstatistics}) and our random partitioning, trivially labelling every example as the majority class gives F1 scores of 0.818 and 0.642 for Subtask A in English and Portuguese, respectively. Although the best English model outperforms this, the best Portuguese model actually performs worse than this trivial baseline. 
This shows there is still significant room for improvement in this task. 
The most interesting result was that models perform surprisingly well in the one-shot and few-shot setups. This is a novel observation, made possible by the unique nature of this dataset and is likely to be very helpful in developing methods of identifying idiomatic language.

We found that including context sentences did not always lead to significantly improved model performance. Intuitively, one would expect an increase in performance due to the availability of more relevant data. A possible reason we did not observe this in our experiments is that we included context by simply concatenating the three sentences, which means the model has no awareness of which sentence is relevant and could be deceived by surrounding sentences containing idiomatic expressions, for example. However, in the zero shot setting, including the context while excluding the target MWE led to a significant increase in generalisability as measured by the increased performance on the test set. This combination led to an increase of almost 8 points over the development set in English and 2 points in Portuguese where all other combinations led to a drop in performance on the test set as compared to the development set. 

The inclusion of the relevant MWE, was generally found to be greatly beneficial to model performance. The intuition behind this is that models are able to ``focus'' on the relevant MWE when determining idiomaticity. In the one and few shot settings in particular, this inclusion significantly boosted performance. When models had previously not encountered examples associated with a particular MWE in the training data (as in the zero shot setting), including the MWE did less to boost performance, although it still did improve results. The only advantage of excluding the MWE was in helping with generalisation as detailed above. In the case of both MWE and context inclusion, we expect more sophisticated methods of incorporating this information to further boost performance.

We note that results in English outperform those in Portuguese. We believe that this difference could be a result of three factors: a) the fact that there is less training data available in Portuguese, b) because models are pre-trained on significantly less Portuguese data, and c) due to the higher degree of inflection in Portuguese. 

\subsection{Representation of Idiomaticity}

Recall that the evaluation data for Task 2 included data from standard STS datasets to ensure that the task is not trivially solvable (Section \ref{section:task2-intro}). We report results on only the MWE subset of the evaluation data in Tables \ref{table:task2-a-erroranalysis} and \ref{table:task2-b-erroranalysis} for Subtasks A and B respectively.

\begin{table}[ht]
\footnotesize
\def\arraystretch{1.1}
\centering
\begin{tabular}{|L{2.5cm}|L{1.5cm}|L{1.5cm}|}
\hline
Tokenization & EN Non-STS $\rho$ & PT Non-STS $\rho$ \\
\hline
Default & 0.219 & 0.203 \\
All Tokenized (No Pre-Training) & 0.395 & 0.274 \\
All Tokenized  & 0.459 & 0.369 \\
Select Tokenized  & 0.437 & 0.332 \\
\hline
\end{tabular}
\caption{\label{table:task2-a-erroranalysis} Results on only the MWE subset of the Test split for Task 2 Subtask A.}
\end{table}

\begin{table}[ht]
\footnotesize
\def\arraystretch{1.1}
\centering
\begin{tabular}{|L{2.5cm}|L{1.5cm}|L{1.5cm}|}
\hline
Tokenization & EN Non-STS $\rho$ & PT Non-STS $\rho$ \\
\hline
Default & 0.627 & 0.312 \\
All Tokenized  & 0.611 & 0.379 \\
Select Tokenized  & 0.618 & 0.416 \\
\hline
\end{tabular}
\caption{\label{table:task2-b-erroranalysis}Results on only the MWE subset of the Test split for Task 2 Subtask B.}
\end{table}

These results show the significant scope for improvement in representing idiomaticity (given model performance on the standard STS benchmark datasets is close 0.9$\rho$). Additionally, we note that in Subtask A, which requires the use of only pre-training, it is better to tokenize all pre-training data, thus maximising the amount of training data, rather than selectively tokenizing training data. In Subtask B (fine-tuning), however, selective tokenization seems to have a slight advantage although the default tokenization seems to be more suitable in English.

Thus, our experiments exploring the use of the idiomatic principle to capture idiomaticity in contextual pre-trained models (Task 2 Subtask A), show that while replacing potential MWEs with a single token does improve performance, further pre-training with text tokenized either using ``all replace'' or ``select replace'' improves performance only on the MWE subset of the evaluation split. On the full test set, which includes standard STS data (Table \ref{table:task2-a-results}), however, additional pre-training with MWE data does not always improve over a random representation of MWE tokens and when it does, it does so only slightly. This is an interesting result, and could be because gains made by the use of the idiom principle are offset by the continuing to pre-train on a relatively small set of sentences that include a randomly initialised token added to the vocabulary, or because the gains made on the MWE subset are diluted across the entire test split. 
%This seems to suggest some of the gains observed by using the idiomatic principle come from ``breaking'' compositionality and not because of the new representations learned for these tokens, possibly due to limited usage of MWEs in natural text, leading to limited pre-training data. This is also supported by the fact that infrequent words are not effectively represented in pre-trained language models~\cite{schick2020rare}. %Our experiments also suggest that ``select replacement'', which involves the use of idiomaticity detection to tokenize input to models creating representations of sentences containing idioms, is a method with significant potential, especially given that this method lends itself to end-to-end models. 

Experiments using fine-tuning (Task 2 Subtask B, see Section \ref{section:experiments-results-task2}, Table \ref{table:task2-a-results}) show, unsurprisingly, that  pre-trained language models are extremely effective in transfer learning. What is particularly interesting, though, is that starting with random embeddings for tokens representing MWEs can lead to comparative (and in some cases slightly better) scores. This suggests that these tokens have at least a reasonable representation level, thus providing a sample efficient method of learning embeddings for them. However, further experiments on different tasks are required to test the extent to which these tokens have been trained. 

\section{Conclusions and Future Work}
\label{section:conclusions-and-future-work}
In this work we presented a novel dataset of naturally occurring idiomatic MWE usage in English and Portuguese, with associated tasks aimed at testing the ability of language models to deal with idiomaticity. In addition, we ran a number of experiments on these tasks. % using transformer-based models. 

In terms of idiomaticity detection, the results of our experiments show these models achieve reasonable performance in the one-shot and few-shot settings, but particularly struggle with the zero-shot setting, where the models encounter unseen MWEs at inference time. 

When it comes to the representation of idiomaticity, our experiments show that while the use of the idiom principle does help in representing MWEs, these gains do not transfer to a significant overall increase in performance on the entire test split. The large number of MWEs makes including all of them in the vocabulary impractical, likewise selectively training models with MWEs of interest is impractical due to the cost of pre-training. This underscores the need for a more nuanced approach to incorporating the idiom principle with pre-trained language models.  Additionally, in creating representations for MWEs that are partially compositional, methods that make use of the representations of constituent words such as attentive mimicking~\cite{schick-schutze-2019-attentive} might be beneficial and we intend to experiment with these methods in future. We also find that pre-training is potentially an effective way of learning these representations, although more experiments are required to test these representations.

There are many avenues for future work using the data presented here, including running cross-lingual experiments across different scenarios of data availability. Although our experiments have been limited to the use of transformer based pre-trained language models, the dataset and tasks we present can be used with any language model. While this work provides a useful dataset for the investigation of idiomaticity, we intend to expand this dataset in order to cover a broader set of languages, and include a wider range of idiomatic MWE types, including more syntactically flexible expressions. One limitation of the dataset is that the paraphrases generated are syntactically rigid, and for Task 2 the replacement sentences may not always be grammatically correct (see ~\autoref{table:subtaskc-v2}). Although this is sufficient for current purposes, future datasets could generate paraphrases per sentence rather than per MWE. 

\subsection*{Acknowledgements}

This work was partially supported by the UK EPSRC grant EP/T02450X/1 and the CDT in Speech and Language Technologies and their Applications funded by UKRI (grant number EP/S023062/1).

\bibliography{anthology,custom}

\begin{thebibliography}{40}
\expandafter\ifx\csname natexlab\endcsname\relax\def\natexlab#1{#1}\fi

\bibitem[{Baldwin et~al.(2003)Baldwin, Bannard, Tanaka, and
  Widdows}]{baldwin-etal-2003-empirical}
Timothy Baldwin, Colin Bannard, Takaaki Tanaka, and Dominic Widdows. 2003.
\newblock \href {https://doi.org/10.3115/1119282.1119294} {An empirical model
  of multiword expression decomposability}.
\newblock In \emph{Proceedings of the {ACL} 2003 Workshop on Multiword
  Expressions: Analysis, Acquisition and Treatment}, pages 89--96, Sapporo,
  Japan. Association for Computational Linguistics.

\bibitem[{Baldwin and
  Villavicencio(2002)}]{baldwin-villavicencio-2002-extracting}
Timothy Baldwin and Aline Villavicencio. 2002.
\newblock \href {https://www.aclweb.org/anthology/W02-2001} {Extracting the
  unextractable: A case study on verb-particles}.
\newblock In \emph{{COLING}-02: The 6th Conference on Natural Language Learning
  2002 ({C}o{NLL}-2002)}.

\bibitem[{Biemann and
  Giesbrecht(2011)}]{biemann-giesbrecht-2011-distributional}
Chris Biemann and Eugenie Giesbrecht. 2011.
\newblock \href {https://www.aclweb.org/anthology/W11-1304} {Distributional
  semantics and compositionality 2011: Shared task description and results}.
\newblock In \emph{Proceedings of the Workshop on Distributional Semantics and
  Compositionality}, pages 21--28, Portland, Oregon, USA. Association for
  Computational Linguistics.

\bibitem[{Cer et~al.(2017)Cer, Diab, Agirre, Lopez-Gazpio, and
  Specia}]{cer-etal-2017-semeval}
Daniel Cer, Mona Diab, Eneko Agirre, I{\~n}igo Lopez-Gazpio, and Lucia Specia.
  2017.
\newblock \href {https://doi.org/10.18653/v1/S17-2001} {{S}em{E}val-2017 task
  1: Semantic textual similarity multilingual and crosslingual focused
  evaluation}.
\newblock In \emph{Proceedings of the 11th International Workshop on Semantic
  Evaluation ({S}em{E}val-2017)}, pages 1--14, Vancouver, Canada. Association
  for Computational Linguistics.

\bibitem[{Constant et~al.(2017)Constant, Eryi{\v{g}}it, Monti, van~der Plas,
  Ramisch, Rosner, and Todirascu}]{constant-etal-2017-survey}
Mathieu Constant, G{\"u}l{\c{s}}en Eryi{\v{g}}it, Johanna Monti, Lonneke
  van~der Plas, Carlos Ramisch, Michael Rosner, and Amalia Todirascu. 2017.
\newblock \href {https://doi.org/10.1162/COLI_a_00302} {{S}urvey: Multiword
  expression processing: A {S}urvey}.
\newblock \emph{Computational Linguistics}, 43(4):837--892.

\bibitem[{Cook et~al.(2008)Cook, Fazly, and Stevenson}]{Cook08thevnctokens}
Paul Cook, Afsaneh Fazly, and Suzanne Stevenson. 2008.
\newblock {The VNCTokens Dataset}.
\newblock In \emph{In proceedings of the MWE workshop. ACL}.

\bibitem[{Cordeiro et~al.(2019)Cordeiro, Villavicencio, Idiart, and
  Ramisch}]{cordeiro-etal-2019-unsupervised}
Silvio Cordeiro, Aline Villavicencio, Marco Idiart, and Carlos Ramisch. 2019.
\newblock \href {https://doi.org/10.1162/coli_a_00341} {Unsupervised
  compositionality prediction of nominal compounds}.
\newblock \emph{Computational Linguistics}, 45(1):1--57.

\bibitem[{Devlin et~al.(2019)Devlin, Chang, Lee, and
  Toutanova}]{devlin-etal-2019-bert}
Jacob Devlin, Ming-Wei Chang, Kenton Lee, and Kristina Toutanova. 2019.
\newblock \href {https://doi.org/10.18653/v1/N19-1423} {{BERT}: Pre-training of
  deep bidirectional transformers for language understanding}.
\newblock In \emph{Proceedings of the 2019 Conference of the North {A}merican
  Chapter of the Association for Computational Linguistics: Human Language
  Technologies, Volume 1 (Long and Short Papers)}, pages 4171--4186,
  Minneapolis, Minnesota. Association for Computational Linguistics.

\bibitem[{Farahmand et~al.(2015)Farahmand, Smith, and
  Nivre}]{farahmand-etal-2015-multiword}
Meghdad Farahmand, Aaron Smith, and Joakim Nivre. 2015.
\newblock \href {https://doi.org/10.3115/v1/W15-0904} {A multiword expression
  data set: Annotating non-compositionality and conventionalization for
  {E}nglish noun compounds}.
\newblock In \emph{Proceedings of the 11th Workshop on Multiword Expressions},
  pages 29--33, Denver, Colorado. Association for Computational Linguistics.

\bibitem[{Garcia et~al.(2021)Garcia, Kramer~Vieira, Scarton, Idiart, and
  Villavicencio}]{garcia-etal-2021-probing}
Marcos Garcia, Tiago Kramer~Vieira, Carolina Scarton, Marco Idiart, and Aline
  Villavicencio. 2021.
\newblock \href {https://www.aclweb.org/anthology/2021.eacl-main.310} {Probing
  for idiomaticity in vector space models}.
\newblock In \emph{Proceedings of the 16th Conference of the European Chapter
  of the Association for Computational Linguistics: Main Volume}, pages
  3551--3564, Online. Association for Computational Linguistics.

\bibitem[{Hashempour and
  Villavicencio(2020)}]{hashempour-villavicencio-2020-leveraging}
Reyhaneh Hashempour and Aline Villavicencio. 2020.
\newblock \href {https://www.aclweb.org/anthology/2020.cogalex-1.9} {Leveraging
  contextual embeddings and idiom principle for detecting idiomaticity in
  potentially idiomatic expressions}.
\newblock In \emph{Proceedings of the Workshop on the Cognitive Aspects of the
  Lexicon}, pages 72--80, Online. Association for Computational Linguistics.

\bibitem[{Hashimoto and Tsuruoka(2016)}]{hashimoto-tsuruoka-2016-adaptive}
Kazuma Hashimoto and Yoshimasa Tsuruoka. 2016.
\newblock \href {https://doi.org/10.18653/v1/P16-1020} {Adaptive joint learning
  of compositional and non-compositional phrase embeddings}.
\newblock In \emph{Proceedings of the 54th Annual Meeting of the Association
  for Computational Linguistics (Volume 1: Long Papers)}, pages 205--215,
  Berlin, Germany. Association for Computational Linguistics.

\bibitem[{Hendrickx et~al.(2013)Hendrickx, Kozareva, Nakov, {\'O}~S{\'e}aghdha,
  Szpakowicz, and Veale}]{hendrickx-etal-2013-semeval}
Iris Hendrickx, Zornitsa Kozareva, Preslav Nakov, Diarmuid {\'O}~S{\'e}aghdha,
  Stan Szpakowicz, and Tony Veale. 2013.
\newblock \href {https://www.aclweb.org/anthology/S13-2025} {{S}em{E}val-2013
  task 4: Free paraphrases of noun compounds}.
\newblock In \emph{Second Joint Conference on Lexical and Computational
  Semantics (*{SEM}), Volume 2: Proceedings of the Seventh International
  Workshop on Semantic Evaluation ({S}em{E}val 2013)}, pages 138--143, Atlanta,
  Georgia, USA. Association for Computational Linguistics.

\bibitem[{Kartsaklis et~al.(2014)Kartsaklis, Kalchbrenner, and
  Sadrzadeh}]{kartsaklis-etal-2014-resolving}
Dimitri Kartsaklis, Nal Kalchbrenner, and Mehrnoosh Sadrzadeh. 2014.
\newblock \href {https://doi.org/10.3115/v1/P14-2035} {Resolving lexical
  ambiguity in tensor regression models of meaning}.
\newblock In \emph{Proceedings of the 52nd Annual Meeting of the Association
  for Computational Linguistics (Volume 2: Short Papers)}, pages 212--217,
  Baltimore, Maryland. Association for Computational Linguistics.

\bibitem[{Katz and Giesbrecht(2006)}]{katz-giesbrecht-2006-automatic}
Graham Katz and Eugenie Giesbrecht. 2006.
\newblock \href {https://www.aclweb.org/anthology/W06-1203} {Automatic
  identification of non-compositional multi-word expressions using latent
  semantic analysis}.
\newblock In \emph{Proceedings of the Workshop on Multiword Expressions:
  Identifying and Exploiting Underlying Properties}, pages 12--19, Sydney,
  Australia. Association for Computational Linguistics.

\bibitem[{Lin(1999)}]{lin-1999-automatic}
Dekang Lin. 1999.
\newblock \href {https://doi.org/10.3115/1034678.1034730} {Automatic
  identification of non-compositional phrases}.
\newblock In \emph{Proceedings of the 37th Annual Meeting of the Association
  for Computational Linguistics}, pages 317--324, College Park, Maryland, USA.
  Association for Computational Linguistics.

\bibitem[{Mikolov et~al.(2013{\natexlab{a}})Mikolov, Chen, Corrado, and
  Dean}]{word2vec}
Tomas Mikolov, Kai Chen, Greg~S. Corrado, and Jeffrey Dean. 2013{\natexlab{a}}.
\newblock \href {http://arxiv.org/abs/1301.3781} {Efficient estimation of word
  representations in vector space}.

\bibitem[{Mikolov et~al.(2013{\natexlab{b}})Mikolov, Sutskever, Chen, Corrado,
  and Dean}]{10.5555/2999792.2999959}
Tomas Mikolov, Ilya Sutskever, Kai Chen, Greg Corrado, and Jeffrey Dean.
  2013{\natexlab{b}}.
\newblock Distributed representations of words and phrases and their
  compositionality.
\newblock In \emph{Proceedings of the 26th International Conference on Neural
  Information Processing Systems - Volume 2}, NIPS'13, page 3111–3119, Red
  Hook, NY, USA. Curran Associates Inc.

\bibitem[{Mitchell and Lapata(2010)}]{mitchell2010composition}
Jeff Mitchell and Mirella Lapata. 2010.
\newblock Composition in distributional models of semantics.
\newblock \emph{Cognitive science}, 34(8):1388--1429.

\bibitem[{Nandakumar et~al.(2019)Nandakumar, Baldwin, and
  Salehi}]{nandakumar-etal-2019-well}
Navnita Nandakumar, Timothy Baldwin, and Bahar Salehi. 2019.
\newblock \href {https://doi.org/10.18653/v1/W19-2004} {How well do embedding
  models capture non-compositionality? a view from multiword expressions}.
\newblock In \emph{Proceedings of the 3rd Workshop on Evaluating Vector Space
  Representations for {NLP}}, pages 27--34, Minneapolis, USA. Association for
  Computational Linguistics.

\bibitem[{Raffel et~al.(2019)Raffel, Shazeer, Roberts, Lee, Narang, Matena,
  Zhou, Li, and Liu}]{DBLP:journals/corr/abs-1910-10683}
Colin Raffel, Noam Shazeer, Adam Roberts, Katherine Lee, Sharan Narang, Michael
  Matena, Yanqi Zhou, Wei Li, and Peter~J. Liu. 2019.
\newblock \href {http://arxiv.org/abs/1910.10683} {Exploring the limits of
  transfer learning with a unified text-to-text transformer}.
\newblock \emph{CoRR}, abs/1910.10683.

\bibitem[{Real et~al.(2020)Real, Fonseca, and Oliveira}]{real2020assin}
Livy Real, Erick Fonseca, and Hugo~Goncalo Oliveira. 2020.
\newblock The assin 2 shared task: a quick overview.
\newblock In \emph{International Conference on Computational Processing of the
  Portuguese Language}, pages 406--412. Springer.

\bibitem[{Reddy et~al.(2011)Reddy, McCarthy, and
  Manandhar}]{reddy-etal-2011-empirical}
Siva Reddy, Diana McCarthy, and Suresh Manandhar. 2011.
\newblock \href {https://www.aclweb.org/anthology/I11-1024} {An empirical study
  on compositionality in compound nouns}.
\newblock In \emph{Proceedings of 5th International Joint Conference on Natural
  Language Processing}, pages 210--218, Chiang Mai, Thailand. Asian Federation
  of Natural Language Processing.

\bibitem[{Reimers et~al.(2016)Reimers, Beyer, and
  Gurevych}]{reimers-etal-2016-task}
Nils Reimers, Philip Beyer, and Iryna Gurevych. 2016.
\newblock \href {https://www.aclweb.org/anthology/C16-1009} {Task-oriented
  intrinsic evaluation of semantic textual similarity}.
\newblock In \emph{Proceedings of {COLING} 2016, the 26th International
  Conference on Computational Linguistics: Technical Papers}, pages 87--96,
  Osaka, Japan. The COLING 2016 Organizing Committee.

\bibitem[{Reimers and Gurevych(2019)}]{reimers-2019-sentence-bert}
Nils Reimers and Iryna Gurevych. 2019.
\newblock \href {https://arxiv.org/abs/1908.10084} {Sentence-bert: Sentence
  embeddings using siamese bert-networks}.
\newblock In \emph{Proceedings of the 2019 Conference on Empirical Methods in
  Natural Language Processing}. Association for Computational Linguistics.

\bibitem[{Sag et~al.(2002)Sag, Baldwin, Bond, Copestake, and
  Flickinger}]{10.5555/647344.724004}
Ivan~A. Sag, Timothy Baldwin, Francis Bond, Ann~A. Copestake, and Dan
  Flickinger. 2002.
\newblock Multiword expressions: A pain in the neck for nlp.
\newblock In \emph{Proceedings of the Third International Conference on
  Computational Linguistics and Intelligent Text Processing}, CICLing '02, page
  1–15, Berlin, Heidelberg. Springer-Verlag.

\bibitem[{Savary et~al.(2017)Savary, Ramisch, Cordeiro, Sangati, Vincze,
  QasemiZadeh, Candito, Cap, Giouli, Stoyanova, and
  Doucet}]{savary-etal-2017-parseme}
Agata Savary, Carlos Ramisch, Silvio Cordeiro, Federico Sangati, Veronika
  Vincze, Behrang QasemiZadeh, Marie Candito, Fabienne Cap, Voula Giouli,
  Ivelina Stoyanova, and Antoine Doucet. 2017.
\newblock \href {https://doi.org/10.18653/v1/W17-1704} {The {PARSEME} shared
  task on automatic identification of verbal multiword expressions}.
\newblock In \emph{Proceedings of the 13th Workshop on Multiword Expressions
  ({MWE} 2017)}, pages 31--47, Valencia, Spain. Association for Computational
  Linguistics.

\bibitem[{Schick and Sch{\"u}tze(2019)}]{schick-schutze-2019-attentive}
Timo Schick and Hinrich Sch{\"u}tze. 2019.
\newblock \href {https://doi.org/10.18653/v1/N19-1048} {Attentive mimicking:
  Better word embeddings by attending to informative contexts}.
\newblock In \emph{Proceedings of the 2019 Conference of the North {A}merican
  Chapter of the Association for Computational Linguistics: Human Language
  Technologies, Volume 1 (Long and Short Papers)}, pages 489--494, Minneapolis,
  Minnesota. Association for Computational Linguistics.

\bibitem[{Schneider et~al.(2016)Schneider, Hwang, Srikumar, Green, Suresh,
  Conger, O{'}Gorman, and Palmer}]{schneider-etal-2016-corpus}
Nathan Schneider, Jena~D. Hwang, Vivek Srikumar, Meredith Green, Abhijit
  Suresh, Kathryn Conger, Tim O{'}Gorman, and Martha Palmer. 2016.
\newblock \href {https://doi.org/10.18653/v1/W16-1712} {A corpus of preposition
  supersenses}.
\newblock In \emph{Proceedings of the 10th Linguistic Annotation Workshop held
  in conjunction with {ACL} 2016 ({LAW}-X 2016)}, pages 99--109, Berlin,
  Germany. Association for Computational Linguistics.

\bibitem[{Schneider et~al.(2014)Schneider, Onuffer, Kazour, Danchik,
  Mordowanec, Conrad, and Smith}]{schneider-etal-2014-comprehensive}
Nathan Schneider, Spencer Onuffer, Nora Kazour, Emily Danchik, Michael~T.
  Mordowanec, Henrietta Conrad, and Noah~A. Smith. 2014.
\newblock \href
  {http://www.lrec-conf.org/proceedings/lrec2014/pdf/521_Paper.pdf}
  {Comprehensive annotation of multiword expressions in a social web corpus}.
\newblock In \emph{Proceedings of the Ninth International Conference on
  Language Resources and Evaluation ({LREC}'14)}, pages 455--461, Reykjavik,
  Iceland. European Language Resources Association (ELRA).

\bibitem[{Schneider and Smith(2015)}]{schneider-smith-2015-corpus}
Nathan Schneider and Noah~A. Smith. 2015.
\newblock \href {https://doi.org/10.3115/v1/N15-1177} {A corpus and model
  integrating multiword expressions and supersenses}.
\newblock In \emph{Proceedings of the 2015 Conference of the North {A}merican
  Chapter of the Association for Computational Linguistics: Human Language
  Technologies}, pages 1537--1547, Denver, Colorado. Association for
  Computational Linguistics.

\bibitem[{Schroff et~al.(2015)Schroff, Kalenichenko, and Philbin}]{7298682}
Florian Schroff, Dmitry Kalenichenko, and James Philbin. 2015.
\newblock \href {https://doi.org/10.1109/CVPR.2015.7298682} {Facenet: A unified
  embedding for face recognition and clustering}.
\newblock In \emph{2015 IEEE Conference on Computer Vision and Pattern
  Recognition (CVPR)}, pages 815--823.

\bibitem[{Schulte~im Walde et~al.(2016)Schulte~im Walde, H{\"a}tty, Bott, and
  Khvtisavrishvili}]{schulte-im-walde-etal-2016-ghost}
Sabine Schulte~im Walde, Anna H{\"a}tty, Stefan Bott, and Nana
  Khvtisavrishvili. 2016.
\newblock \href {https://www.aclweb.org/anthology/L16-1362} {{G}ho{S}t-{NN}: A
  representative gold standard of {G}erman noun-noun compounds}.
\newblock In \emph{Proceedings of the Tenth International Conference on
  Language Resources and Evaluation ({LREC}'16)}, pages 2285--2292,
  Portoro{\v{z}}, Slovenia. European Language Resources Association (ELRA).

\bibitem[{Shwartz and Dagan(2019)}]{shwartz-dagan-2019-still}
Vered Shwartz and Ido Dagan. 2019.
\newblock \href {https://doi.org/10.1162/tacl_a_00277} {Still a pain in the
  neck: Evaluating text representations on lexical composition}.
\newblock \emph{Transactions of the Association for Computational Linguistics},
  7:403--419.

\bibitem[{Sinclair et~al.(1991)Sinclair, Sinclair, and
  Carter}]{sinclair1991corpus}
J.~Sinclair, L.~Sinclair, and R.~Carter. 1991.
\newblock \href {https://books.google.co.uk/books?id=L8l4AAAAIAAJ}
  {\emph{Corpus, Concordance, Collocation}}.
\newblock Describing English language. Oxford University Press.

\bibitem[{Tu and Roth(2012)}]{tu-roth-2012-sorting}
Yuancheng Tu and Dan Roth. 2012.
\newblock \href {https://www.aclweb.org/anthology/S12-1010} {Sorting out the
  most confusing {E}nglish phrasal verbs}.
\newblock In \emph{*{SEM} 2012: The First Joint Conference on Lexical and
  Computational Semantics {--} Volume 1: Proceedings of the main conference and
  the shared task, and Volume 2: Proceedings of the Sixth International
  Workshop on Semantic Evaluation ({S}em{E}val 2012)}, pages 65--69,
  Montr{\'e}al, Canada. Association for Computational Linguistics.

\bibitem[{Venkatapathy and Joshi(2005)}]{venkatapathy-joshi-2005-measuring}
Sriram Venkatapathy and Aravind Joshi. 2005.
\newblock \href {https://www.aclweb.org/anthology/H05-1113} {Measuring the
  relative compositionality of verb-noun ({V}-n) collocations by integrating
  features}.
\newblock In \emph{Proceedings of Human Language Technology Conference and
  Conference on Empirical Methods in Natural Language Processing}, pages
  899--906, Vancouver, British Columbia, Canada. Association for Computational
  Linguistics.

\bibitem[{Yang et~al.(2019)Yang, Dai, Yang, Carbonell, Salakhutdinov, and
  Le}]{NEURIPS2019_dc6a7e65}
Zhilin Yang, Zihang Dai, Yiming Yang, Jaime Carbonell, Russ~R Salakhutdinov,
  and Quoc~V Le. 2019.
\newblock \href
  {https://proceedings.neurips.cc/paper/2019/file/dc6a7e655d7e5840e66733e9ee67cc69-Paper.pdf}
  {Xlnet: Generalized autoregressive pretraining for language understanding}.
\newblock In \emph{Advances in Neural Information Processing Systems},
  volume~32. Curran Associates, Inc.

\bibitem[{Yu and Ettinger(2020)}]{yu-ettinger-2020-assessing}
Lang Yu and Allyson Ettinger. 2020.
\newblock \href {https://doi.org/10.18653/v1/2020.emnlp-main.397} {Assessing
  phrasal representation and composition in transformers}.
\newblock In \emph{Proceedings of the 2020 Conference on Empirical Methods in
  Natural Language Processing (EMNLP)}, pages 4896--4907, Online. Association
  for Computational Linguistics.

\bibitem[{Zhang et~al.(2019)Zhang, Han, Liu, Jiang, Sun, and
  Liu}]{zhang-etal-2019-ernie}
Zhengyan Zhang, Xu~Han, Zhiyuan Liu, Xin Jiang, Maosong Sun, and Qun Liu. 2019.
\newblock \href {https://doi.org/10.18653/v1/P19-1139} {{ERNIE}: Enhanced
  language representation with informative entities}.
\newblock In \emph{Proceedings of the 57th Annual Meeting of the Association
  for Computational Linguistics}, pages 1441--1451, Florence, Italy.
  Association for Computational Linguistics.

\end{thebibliography}
\bibliographystyle{acl_natbib}

\clearpage
\appendix

\section{Dataset Statistics}
\label{app:app-datasetstatistics}

Detailed statistics for the English and Portuguese datasets are shown in \autoref{table:enDatasetBreakdown} and \autoref{table:ptDatasetBreakdown}, respectively. The train, dev and set breakdowns are shown in the leftmost column, with the further breakdown of the train set into zero-shot, one-shot and few-shot setups. Note here that the one-shot data is contained within the few-shot data. The MWEs column is the number of MWEs that the examples span - the one-shot and few-shot setups contain all the MWEs from the dev and test sets, but the examples are different. The next columns show the fine-grained and coarse-grained breakdown of the dataset, used in Task 1 Subtask B and Task 1 Subtask A, respectively.

\begin{table*}[ht]
\centering
\begin{tabular}{ |l l|r |r r r | r r r r r | r| } 
\hline
Set& &  & \multicolumn{3}{c|}{Non-Idiomatic (1)} & \multicolumn{5}{c|}{Idiomatic (0)} & \\
& &  MWEs & Lit & PN & Tot & 1 & 2 & 3 & Meta & Tot & Tot \\
\hline
\multirow{4}{*}{train} & zero-shot & 163 & 1110 & 455 & 1565 & 1614 & 92 & 8 & 48 & 1762 & 3327  \\ 
& (one-shot) & 60 & 29 & 26 & 55 & 25 & 5 & 0 & 2 & 32 & 87  \\ 
& few-shot & 60 & 135 & 50 & 185 & 81 & 11 & 0 & 5 & 97 & 282  \\ 
\cline{2-12}
& total & 223 & 1245 & 505 & 1750 & 1695 & 103 & 8 & 53 & 1859 & 3609\\ 
\hline
dev & & 30 & 174 & 110 & 284 & 157 & 14 & 0 & 11 & 182 & 466\\ 
test & & 30 & 271 & 63 & 334 & 118 & 24 & 0 & 7 & 149 & 483\\ 
\hline
total & & 223 & 1690 & 678 & 2368 & 1970 & 141 & 8 & 71 & 2190 & 4558\\
\hline
\end{tabular}
\caption{Breakdown of the English dataset.}
\label{table:enDatasetBreakdown}
\end{table*}

\begin{table*}[ht]
\centering
\begin{tabular}{ |l l|r |r r r |r r r r r | r| } 
\hline
Set& &  & \multicolumn{3}{c|}{Non-Idiomatic (1)} & \multicolumn{5}{c|}{Idiomatic (0)} & \\
& &  MWEs & Lit & PN & Tot & 1 & 2 & 3 & Meta & Tot & Tot \\
\hline
\multirow{4}{*}{train} & zero-shot & 73 & 284 & 107 & 391 & 697 & 55 & 2 & 19 & 773 & 1164 \\ 
& (one-shot) & 40 & 17 & 8 & 25 & 26 & 2 & 0 & 0 & 28 & 53 \\ 
& few-shot & 40 & 55 & 14 & 69 & 80 & 6 & 0 & 1 & 87 & 156 \\ 
\cline{2-12}
& total & 113 & 339 & 121 & 460 & 777 & 61 & 2 & 20 & 860 & 1320\\ 
\hline
dev & & 20 & 96 & 23 & 119 & 137 & 16 & 0 & 1 & 154 & 273 \\ 
test & & 20 & 94 & 20 & 114 & 151 & 9 & 0 & 5 & 165 & 279 \\ 
\hline
total & & 113 & 529 & 164 & 693 & 1065 & 86 & 2 & 26 & 1179 & 1872\\
\hline
\end{tabular}
\caption{Breakdown of the Portuguese dataset.}
\label{table:ptDatasetBreakdown}
\end{table*}

\section{Experimental Procedure}
\label{app:app-experimentalProcedure}
\subsection{Task 1 Subtask A}
The Task 1 experiments were run on NVIDIA Tesla K80s.
For the first subtask, we ran a range of experiments, varying the model used, whether context was used, and whether the MWE was used.
We ran each experiment for 9 epochs with five seeds (0 - 5). 
For all experiments, we used a max sequence length of 128 and a learning rate of 2e-5. The standard tokenizers for each model were used for tokenizing the input. 
The results for the best performing seed and epoch (by F1 score) for each experiment are shown in ~\autoref{table:devResultsSubtask1a}, with approximate training run times (for one seed for nine epochs). We started with the zero-shot experiments, then took the best performing models, and continued training them from the best epoch for another 9 epochs in the one-shot and few-shot setups.

\subsection{Task 1 Subtask B}
For the second subtask, we took the best-performing experimental settings from Subtask A: XLNET base (cased) for English and BERT base multilingual (cased) for Portuguese, excluding context but not including the MWE since we instead pass the relevant paraphrase of the MWE (either correct or incorrect). These models were then trained on the multiclass data, again for 9 epochs and with five seeds (0-5). Again, the best-performing models in the zero-shot setup were continued training from the best epoch for another 9 epochs in the few-shot and one-shot setups. These experiments are shown in ~\autoref{table:devResultsSubtask1b}. Training times are increased due to the larger dataset from generation of negative samples.

\begin{table*}[ht]
\centering
\footnotesize
\begin{tabular}{ | c | c|c c c  | c | c c |} 
\hline
& Problem Setup & Model & Context? & MWE? & Train Time & Dev Accuracy & Dev F1 \\
\hline
\multirow{6}{*}{\rotatebox{90}{English~~~~~~~~~~~~~~~~~~~~~~~}} & \multirow{4}{*}{zero-shot} & BERT base (cased) & No & No & \textasciitilde 1 hour & 0.732 & 0.724 \\ 
&& BERT base (cased) & Yes & No & \textasciitilde 1 hour & 0.732 & 0.717 \\ 
&& BERT base (cased) & Yes & Yes & \textasciitilde 1 hour & 0.785 & 0.779\\ 
&& BERT base (cased) & No & Yes & \textasciitilde 1 hour & 0.796 & 0.785 \\ 
&& BERT base (uncased) & No & Yes & \textasciitilde 1 hour & 0.777 & 0.77 \\ 
&& XLNet base (cased) & No & Yes & \textasciitilde 1 hour & 0.828 & 0.823\\ 
&& DistilBERT base (cased) & No & Yes & \textasciitilde 1 hour & 0.768 & 0.757 \\ 
&& RoBERTa base (cased) & No & Yes & \textasciitilde 1 hour & 0.807 & 0.801 \\ 
\cline{2-8}
& one-shot & XLNet base (cased) & No & Yes &  +\textasciitilde5 mins & 0.903 & 0.897\\ 
& one-shot & XLNet base (cased) & Yes & No &   +\textasciitilde5 mins & 0.719 & 0.689\\ 
& one-shot & XLNet base (cased) & No & No &   +\textasciitilde5 mins & 0.775 & 0.755\\
\cline{2-8}
&few-shot & XLNet base (cased) & No & Yes &  +\textasciitilde1min & 0.961 & 0.959\\ 
&few-shot & XLNet base (cased) & Yes & No &  +\textasciitilde1min & 0.807 & 0.782\\
&few-shot & XLNet base (cased) & No  & No &  +\textasciitilde1min & 0.813 & 0.792\\
\hline
\multirow{6}{*}{\rotatebox{90}{Portuguese~~~~~~~~~~~~~~~~}} & \multirow{4}{*}{zero-shot} & XLM-RoBERTa base (cased) & No & No & \textasciitilde 1 hour & 0.604 & 0.593 \\ 
 && XLM-RoBERTa base (cased) & Yes & No & \textasciitilde 1 hour & 0.56 & 0.542\\
 && XLM-RoBERTa base (cased) & Yes & Yes & \textasciitilde 1 hour & 0.714 & 0.696\\
 && XLM-RoBERTa base (cased) & No & Yes & \textasciitilde 1 hour & 0.729 & 0.703\\ 
 && BERT base multilingual (cased)& No & Yes & \textasciitilde 1 hour & 0.707 & 0.686\\ 
\cline{2-8}
& one-shot & XLM-RoBERTa base (cased) & No & Yes &   +\textasciitilde5 mins & 0.879 & 0.877\\ 
& one-shot & XLM-RoBERTa base (cased) & Yes & No &   +\textasciitilde5 mins & 0.615 & 0.605\\
& one-shot & XLM-RoBERTa base (cased) & No  & No &   +\textasciitilde5 mins & 0.641 & 0.638\\
\cline{2-8}
&few-shot & XLM-RoBERTa base (cased) & No & Yes &  +\textasciitilde1min & 0.927 & 0.926 \\
&few-shot & XLM-RoBERTa base (cased) & Yes & No &  +\textasciitilde1min & 0.656 & 0.655\\
&few-shot & XLM-RoBERTa base (cased) & No  & No &  +\textasciitilde1min & 0.799 & 0.796\\
\hline
\end{tabular}
\caption{\label{table:devResultsSubtask1a}Dev set results for Task 1 Subtask A }
\end{table*}

\begin{table*}[ht]
\centering
\footnotesize
\begin{tabular}{ | c | c | c | c | c c |} 
\hline
& Problem Setup & Model &Train Time & Dev Accuracy & Dev F1\\
\hline
\multirow{3}{*}{\rotatebox{90}{En}} & zero-shot & XLNet base (cased) &  \textasciitilde 2.5 hours & 0.883 & 0.852 \\ 
\cline{2-6}
& one-shot & XLNet base (cased) &   +\textasciitilde 20 mins &  0.938 & 0.923 \\ 
\cline{2-6}
&few-shot & XLNet base (cased) &   +\textasciitilde 1 hour & 0.947 & 0.933 \\ 
\hline
\multirow{3}{*}{\rotatebox{90}{Pt}} & zero-shot & XLM-RoBERTa base (cased) &   \textasciitilde 1 hour & 0.886 & 0.843 \\ 
\cline{2-6}
& one-shot & XLM-RoBERTa base (cased) &   +\textasciitilde 5 mins & 0.888 & 0.852  \\ 
\cline{2-6}
& few-shot & XLM-RoBERTa base (cased) & +\textasciitilde 20 mins & 0.931 & 0.909 \\ 
\hline
\end{tabular}
\caption{\label{table:devResultsSubtask1b}Dev set results for Task 1 Subtask B}
\end{table*}

\subsection{Task 2}
Pre-training models was done using NVIDIA Tesla V100s and took approximately 15 hours for each of the two models in English (BERT base on ``all replaced'' and ``select replaced'') and 5 hours for each model in Portuguese (BERT base multilingual). Due to time and resource limitations, we pre-train models only once. All models were pre-trained for 5 epochs based on the evaluation on a development set and our initial experiments which showed that further pre-training did not improve results.

For Subtask A, fine-tuning these models using the Sentence BERT architecture (so as to be able to compare the resultant embeddings using cosine similarity) was done using NVIDIA K80 GPUs for English and took approximately 6 minutes per seed. Since we tested four variations (original BERT, BERT tokenized but not pre-trained, BERT all tokenized and select tokenized) each with five seeds, these experiments took a total of about two hours. The multilingual models required the use of NVIDIA Tesla V100s due to their larger size and took about 3 minutes to train each model (per seed) and consequently took a total of about an hour to train. The best model was picked based on the performance on the STS dataset they were trained on (i.e. the STS benchmark dataset for English and ASSIN2 for Portuguese).

Subtask B similarly required the use of NVIDIA K80 GPUs for English and NVIDIA Tesla V100s for Portuguese. We select the best models fine-tuned using the Sentence BERT architecture (with no pre-training) from Subtask A and continue pre-training with MWE specific data. This process took approximately 6 minutes per model in English and 3 minutes in Portuguese leading to a total of about 30 minutes and 15 minutes respectively. 

All fine-tuning was done for as many epochs as was required to see a drop in performance on the corresponding development set. %This was 2, 2, and 3 epochs for the default tokenization, All tokenization and Select tokenization respectively in English and 1, 5, and 5 epochs for the default tokenization, All tokenization and Select tokenization respectively in Portuguese. 

\subsection{Larger Models}
Exploratory experiments on Task 1 showed that the larger language models performed worse than the base ones, and thus these were the ones we used in our experiments. % We are uncertain as to the reason for this, and...

For task 2, we use the smaller base models due to the limited amount of pre-train data, which we believe would make the use of larger models impractical. 
\end{document}